# Three-dimentional reconstruction of complex, dynamic population canopy architecture for crops with a novel point cloud completion model: A case study in *Brassica napus* rapeseed

Ziyue Guo[a,b], Xin Yang[a,b], Yutao Shen[a,b], Yang Zhu[c], Lixi Jiang[c], Haiyan Cen[a,b,]*

[a] *State Key Laboratory for Vegetation Structure, Function and Construction (VegLab), College of Biosystems* Engineering and *Food Science, Zhejiang University, Hangzhou 310058, P.R. China*

[b] *Key Laboratory of Spectroscopy Sensing, Ministry of Agriculture and Rural Affairs, Hangzhou 310058, P.R. China*

[c] *Institute of Crop Science, Zhejiang University, Hangzhou 310058, P.R. China*

* *Correspondence: hycen@zju.edu.cn; Tel: +86-571-8898-2527*

## Abstract

Quantitative descriptions of the complete canopy architecture are essential for accurately evaluating crop photosynthesis and yield performance to guide ideotype design. Although various sensing technologies have been developed for three-dimensional (3D) reconstruction of individual plants and canopies, they failed to obtain an accurate description of canopy architectures due to severe occlusion among complex canopy architectures. We proposed an effective method for 3D reconstruction of complex, dynamic population canopy architecture for rapeseed crops with a novel point cloud completion model. A complete point cloud generation framework was developed for automated annotation of the training dataset by distinguishing surface points from occluded points within canopies. The crop population point cloud completion network (CP-PCN) was then designed with a multi-resolution dynamic graph convolutional encoder (MRDG) and a point pyramid decoder (PPD) to predict occluded points. To further enhance feature extraction, a dynamic graph convolutional feature extractor (DGCFE) module was proposed to capture structural variations over the whole rapeseed growth period. The results demonstrated that CP-PCN achieved chamfer distance (CD) values of 3.35 cm -4.51 cm over four growth stages, outperforming the state-of-the-art transformer-based method (PoinTr). Ablation studies confirmed the effectiveness of the MRDG and DGCFE modules. Moreover, the validation experiment demonstrated that the silique efficiency index developed from CP-PCN improved the overall accuracy of rapeseed yield prediction by 11.2% compared to that of using incomplete point clouds. The CP-PCN pipeline has the potential to be extended to other crops, significantly advancing the quantitatively analysis of in-field population canopy architectures.

**Introduction**

Development of dense-planting adapted rapeseed (*Brassica napus* L.) cultivars with high yield potential is essential to prevent economic crops from competing with staple food crops for arable land. Rational crop population canopy architecture is a crucial strategy for achieving this goal, as the number, size, and spatial arrangement of canopy leaves directly influence light distribution and interception, thereby affecting photosynthetic efficiency and ultimately impacting yield (Liu et al., 2021). In sparse canopies, horizontal leaves maximize light interception, whereas in dense canopies, an ideal architecture features upright leaves at the top, with progressively horizontal leaves in deeper layers, optimizing light utilization efficiency. Furthermore, the architectural properties of canopy stems contribute significantly to environmental adaptability. Taller stems are more prone to lodging under high winds, while shorter stems may limit leaf area, thereby reducing yield potential (Wang et al., 2024). Additionally, canopy architecture influences the microenvironment within the crop population. Overly dense canopies restrict air circulation, increasing the risk of disease, while sparse canopies fail to meet high-yield requirements (Li et al., 2022). However, quantitative descriptions of canopy architecture that contributes to conducting ideotype design to improve photosynthesis and yield, especially for rapeseed crops with large canopy architecture diversity across the entire growth period, remains a great challenge.

Recently development of high-throughput plant phenotyping with three-dimensional (3D) imaging and reconstruction methods provides an opportunity to characterize crop canopy architecture. These techniques include depth camera, laser scanning, light detection and ranging (LiDAR), and multi-view imaging. Reported studies have used the depth camera in plant phenotyping systems for 3D data acquisition, enabling the extraction of key structural parameters such as the leaf count and the leaf area of tomato plants (Masuda, 2021; Watawana and Isaksson, 2024). Hand-held laser scanner was also utilized to capture fine-scale 3D point clouds of individual rapeseed plants at the maturity stage, allowing the extraction of structural information such as silique volume and length (Ma et al., 2023). While these methods are limited by the imaging conditions required by the sensors and their detection range, making them suitable only for reconstructing simple targets such as individual plants or greenhouse crops. Active LiDAR systems combined with unmanned aerial vehicle (UAV) provide an alternative for acquiring large-scale crop canopy point clouds at the plot or field scale, achieving accurate extraction of maize crop structural parameters such as plant height with the coefficient of determination ($R^2$) > 0.97 and leaf area index of $R^2$ = 0.96 (Bailey and Mahaffee, 2017; Jin et al., 2021). While the high cost and light absorption issues of LiDAR at the emission wavelength limits its applications to obtain detailed features of canopy internal architecture. Thanks to the fast development of deep learning, and multi-view imaging combined with recent emergence of advanced deep learning algorithms such as neural radiance fields (NeRF), has significantly improved the accuracy of 3D crop reconstruction from multi-view images, facilitating successful 3D reconstruction of crops in both greenhouses and field environments (Arshad et al., 2024). With the additional color information and easy to fuse with other sensing data, multi-view imaging-based 3D reconstruction has been widely utilized in the extraction of various canopy structural parameters (Li et al., 2022a; Rossi et al., 2022). However, due to severe occlusion within the canopy, particularly in the late growth stages, it is still unable to capture the complete canopy architecture, especially the internal parts.

To address these challenges, an ideal approach is to quantitatively describe the spatial architecture of plants and then use computer graphics techniques to generate virtual crop models with complete canopy architectures. Murchie and Burgess (2022) analyzed the photosynthetic efficiency of crops under different virtual canopy architectures by modifying leaf distribution, providing a theoretical basis for ideotype plant design. While crop growth is influenced by various environmental factors, making it difficult to precisely quantify the variations in canopy architecture, leading to simulated results that often fail to accurately reflect real-world conditions. Reported studies have attempted to use shape-fitting methods to complete occlusive parts at the plant organ-level. Ge et al. (2020) achieved point cloud completion for occlusive fruits using predefined topological rules. The completed strawberry point cloud demonstrated a high intersection over union (IoU) of 0.77 compared to the ground truth. Lou et al. (2022) used biological constraints to complete occlusive leaf point cloud. Based on the completed lettuce point clouds, the $R^2$ value of the projected leaf area increased from 0.741 to 0.911, while the $R^2$ for total leaf area estimation improved from 0.338 to 0.964. These methods rely on limited prior knowledge and quantitative descriptions of structured objects, making them suitable only for organs with symmetrical architectures.

With the development of deep learning, significant advancements have been made in point cloud completion tasks, primarily driven by large datasets and improvements in network design. Many studies have made their 3D datasets publicly available, from which random clipping can be applied to generate a large number of training samples, providing a solid foundation for point cloud completion models (Geiger et al., 2013; Wu et al., 2015; Tchapmi et al., 2019). To address the unordered and unstructured nature of point clouds, some studies have designed various networks for feature extraction, including point-based multi-layer perceptron (MLP), graph-based structures, transformer-based models, and voxel-based convolutions (Xie et al., 2020; Huang et al., 2020; Yu et al., 2021; Fei et al., 2022). These methods have achieved good completion results on public datasets, most of which consist of rigid objects such as vehicles and furniture. In the agricultural domain, point cloud completion studies have primarily focused on simpler structures, such as fruit or potted plant organs. Magistri et al. (2024a) designed a transformer-based model combined with template matching to complete point clouds of occlusive sweet peppers and strawberries obtained from single-view depth images. Similarly, Chen et al. (2023) successfully applied the point-based PF-Net model to complete leaf point clouds, obtaining complete leaf structures from top-down depth camera images of potted cabbage. Likewise, Zhang et al. (2023) incorporated the MSGRNet+OA module into the convolution-based GRNet to successfully complete the occlusive parts of corn leaves scanned by LiDAR. Existing point cloud completion models are typically designed to extract structural features at fixed scales, which limits their ability to dynamically adapt to the evolving plant architecture across different growth stages, particularly for complex crop canopies with overlapping leaves and spatial heterogeneity.

Current approaches face significant challenges in obtaining accurate, complete point clouds for crop populations throughout their growth period. These challenges stem from a lack of large, comprehensive datasets with fully reconstructed canopy architectures, which are essential for training robust models. Existing methods for dataset generation also fail to adequately represent the occluded internal points in crop canopies, which are crucial for point cloud completion. Additionally, the design of point cloud completion networks capable of handling complex, dense, and irregular canopy architectures remains an unresolved issue. To address these challenges, this study makes several key contributions: we generated a simulated rapeseed population dataset with complete canopy architectures using the virtual-real integration (VRI) method, which combines 3D data from individual plants grown under field conditions and the breeding plot planting pattern. This approach also integrates an occlusion point detection algorithm to create a point cloud completion dataset that accurately reflects real-world conditions. We designed the crop population point cloud completion network (CP-PCN) based on the generative adversarial network (GAN) framework, incorporating a multi-resolution encoder and dynamic graph convolutional feature extractor (DGCFE) to handle the complexity of canopy architectures across different growth stages. Furthermore, we proposed the silique efficiency index (SEI) as a novel metric for estimating yield potential, demonstrating that complete point clouds significantly improve yield prediction accuracy.

# Results

## 3D reconstruction pipeline of complex, dynamic population canopy complete architecture for crops

As shown in Figure 1, we propose a novel 3D reconstruction pipeline designed for dynamic and complex crop canopy complete architecture. The process involves the generation of a richly annotated dataset for training the CP-PCN, which is then used to predict occluded canopy architectures in field-grown crop populations. The dataset generation process consists of three main steps. First, a diverse set of 3D data for individual plants is collected. Second, the complete canopy architecture of a simulated crop population is generated based on planting patterns. Third, an occlusion point detection algorithm is proposed to differentiate between surface points and occluded points, which cannot be reconstructed from multi-view UAV imagery. After training the CP-PCN model with this dataset, it successfully predicts the occluded parts of the canopy when given input surface point clouds, resulting in a complete reconstructed canopy. The completed point cloud is then used for extracting key crop canopy traits, and through a newly proposed architectural index, it enables high-precision yield estimation.

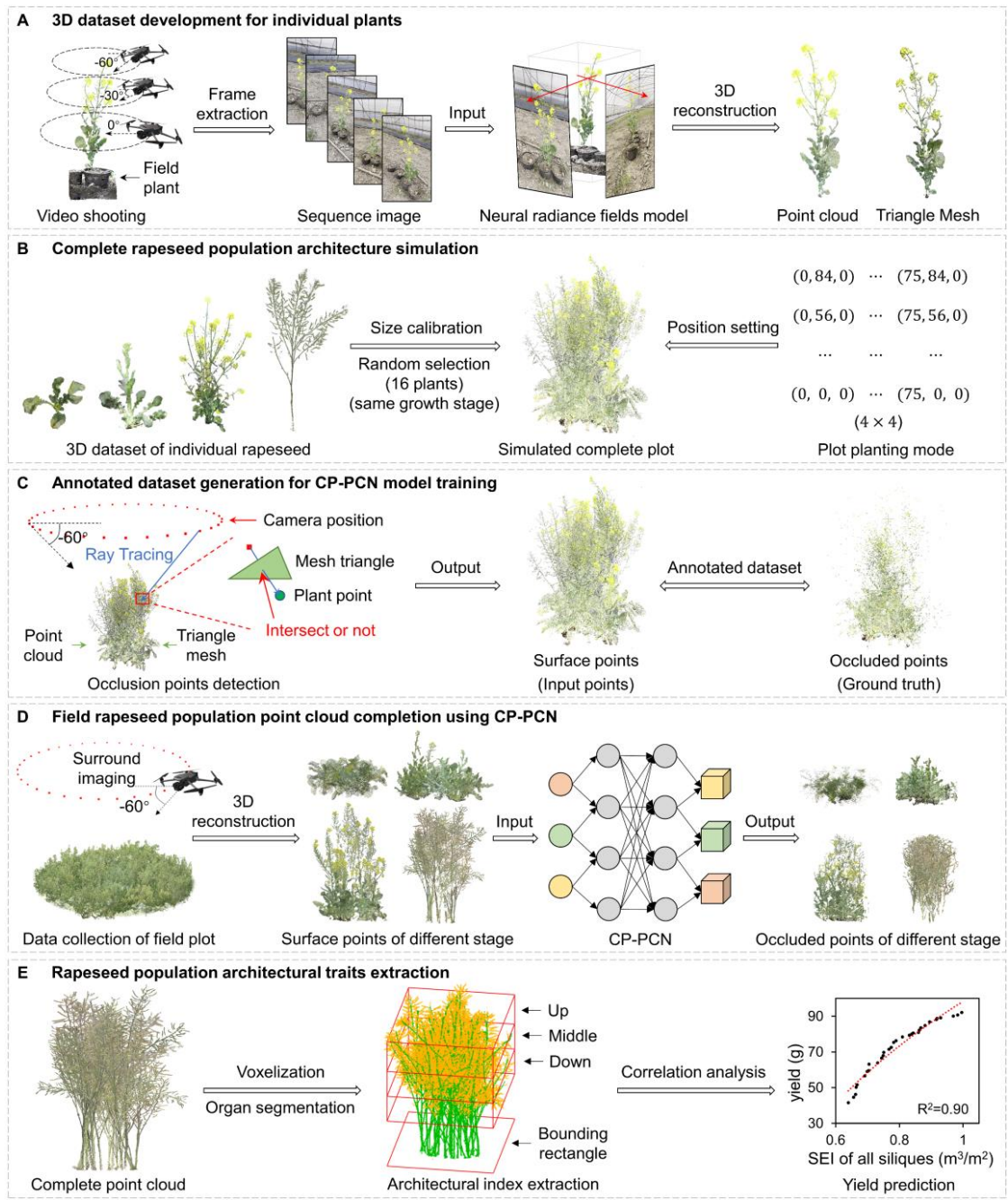

**Figure 1.** 3D reconstruction pipeline of complex, dynamic population canopy complete architecture for crops. (**A**) Development of the 3D dataset for individual rapeseed plants using UAV-based surround filming and neural

radiance fields (NeRF) reconstruction. (**B**) Construction of the simulated population point cloud using the virtual-real integration (VRI) method, which combines the 3D models of field-grown individuals with a realistic planting pattern. (**C**) An occlusion point detection algorithm was developed to distinguish surface points from occluded points, enabling automated annotation of the training dataset. (**D**) Field data collection from rapeseed populations, applying the trained CP-PCN model to complete the occluded parts of the canopy. (**E**) Extraction of crop population architectural traits.

**Structural similarity between simulated and real point clouds of rapeseed populations**

Figure 2A and B present both qualitative and quantitative evaluations of the simulated crop population canopy architecture using three different methods: the proposed VRI method, simple repetition of individual field-grown plants, and simple repetition of individual potted plants. The qualitative results in Figure 2A clearly show that the VRI method closely resembles the actual field-grown rapeseed populations across all growth stages, with particularly accurate representation during the seedling, bolting, and flowering stages. In contrast, the simple repetition of field-grown plants yields a much less accurate simulation, with substantial discrepancies, especially during the silique stage. This method fails to capture the structural complexity and diversity inherent in a natural rapeseed population. The repetition of potted plants exhibits even greater variation from the actual field conditions, with architectures significantly simpler than those observed in field-grown plants, leading to poor simulation results throughout all growth stages.

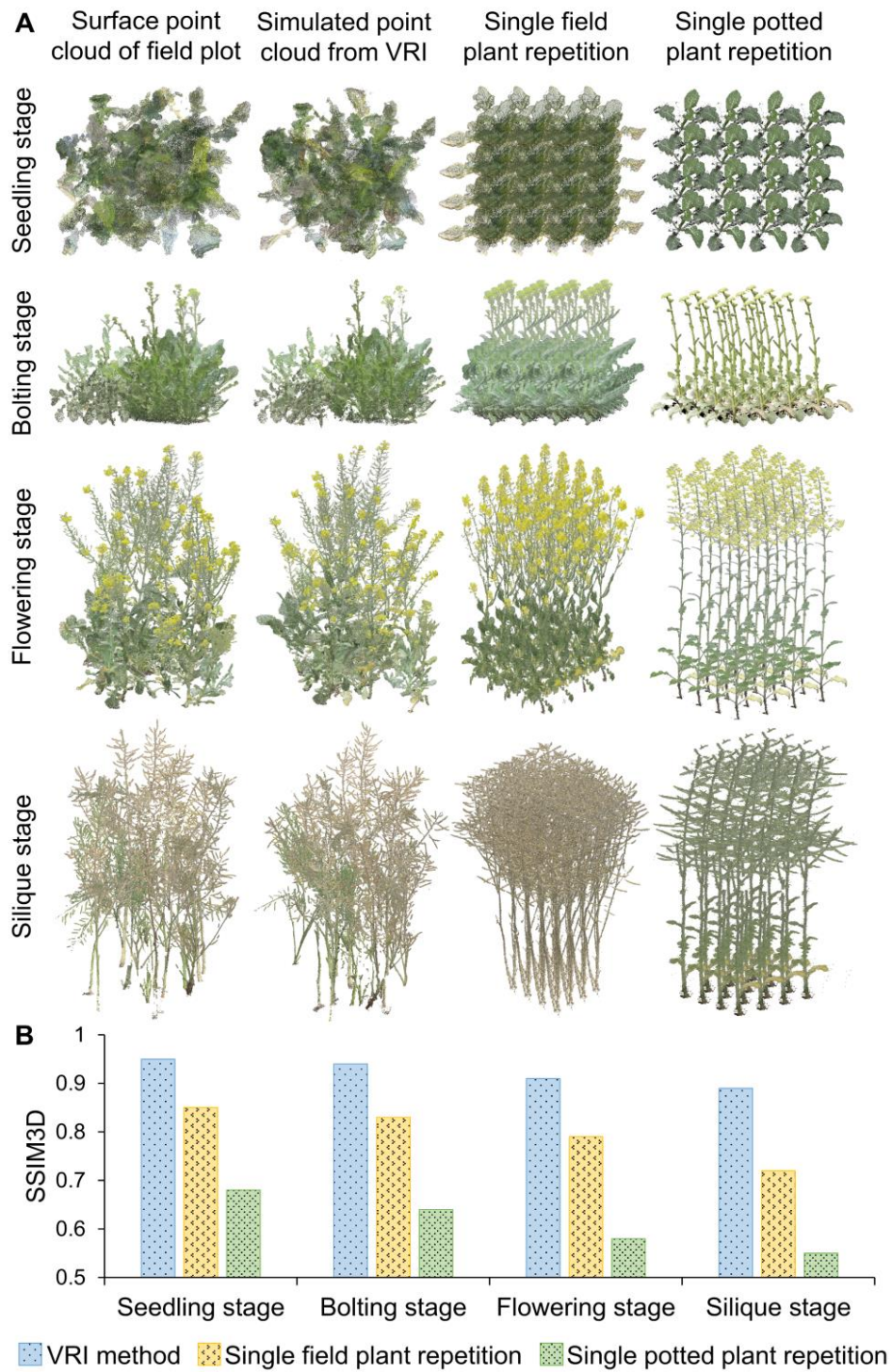

**Figure 2.** Evaluation of simulated crop population canopy architecture. (**A**) Comparison of canopy architectures across four growth stages using different

simulation strategies: VRI method, single field plant repetition, and single potted plant repetition. (**B**) Quantitative comparison of structural similarity (SSIM3D) between simulated and actual rapeseed population point clouds for each strategy across growth stages.

Quantitative results shown in Figure 2B further support these findings. The SSIM3D values for the VRI method were consistently higher than those for the other two methods across all stages. The VRI method achieved the highest SSIM3D values of 0.95, 0.94, 0.91, and 0.89 for the seedling, bolting, flowering, and silique stages, respectively, demonstrating its ability to replicate real canopy architectures under field conditions. In contrast, the repetition of individual field-grown plants produced moderate similarity scores, with SSIM3D values decreasing from 0.85 at the seedling stage to 0.72 at the silique stage. The potted plant repetition method performed the worst, with similarity scores dropping from 0.68 to 0.55 across the same growth stages.

**Evaluation of CP-PCN performance**

The CP-PCN model demonstrated robust performance in reconstructing rapeseed canopy architectures across four growth stages. As illustrated in Figure 3, the model successfully reconstructed internal canopy architectures, with slices along the X, Y, and Z planes highlighting the internal details of the completed point clouds. Enlarged regions further emphasized the model's ability to recover occluded areas. The full point cloud results, providing a comprehensive view of the reconstructed canopy, are provided in

Supplemental Note 1. At the seedling stage, the model effectively reconstructed sparse input data, achieving accurate point cloud generation across all axes. As complexity increased during the bolting, flowering, and silique stages, the model continued to demonstrate high accuracy, handling intricate canopy architectures and recovering 3D architectural details with the minimal deviation.

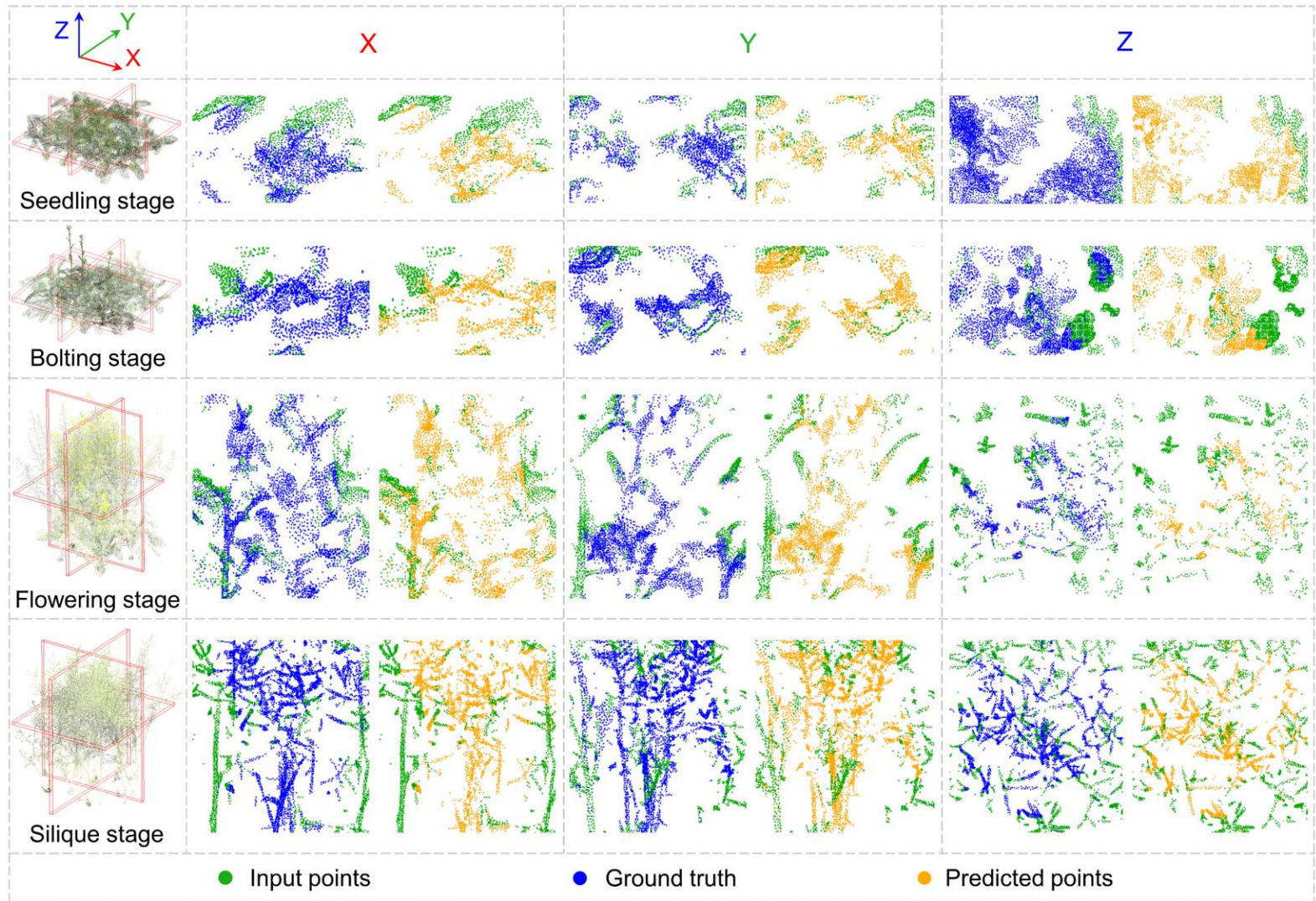

**Figure 3.** Visualization of the complete point cloud at four different growth stages generated by the crop population point cloud completion network (CP-PCN). Examples of the complete point cloud extracted from the X, Y, and Z planes are presented with the input points, ground truth and predicted points denoted by green, blue, and orange dots, respectively.

To evaluate the performance of CP-PCN in real-world scenarios, the model was tested using incomplete point clouds collected from field-grown rapeseed populations. As shown in Figure 4A, a comparison between the incomplete input data and the completed point clouds generated by CP-PCN illustrates the model's robust ability to fill in missing plant architectures, particularly in regions like the silique and branches. The model's ability to address occlusions and restore detailed canopy morphology was clearly demonstrated, with the predicted point clouds (shown in orange) effectively filling the gaps in the original input data (green).

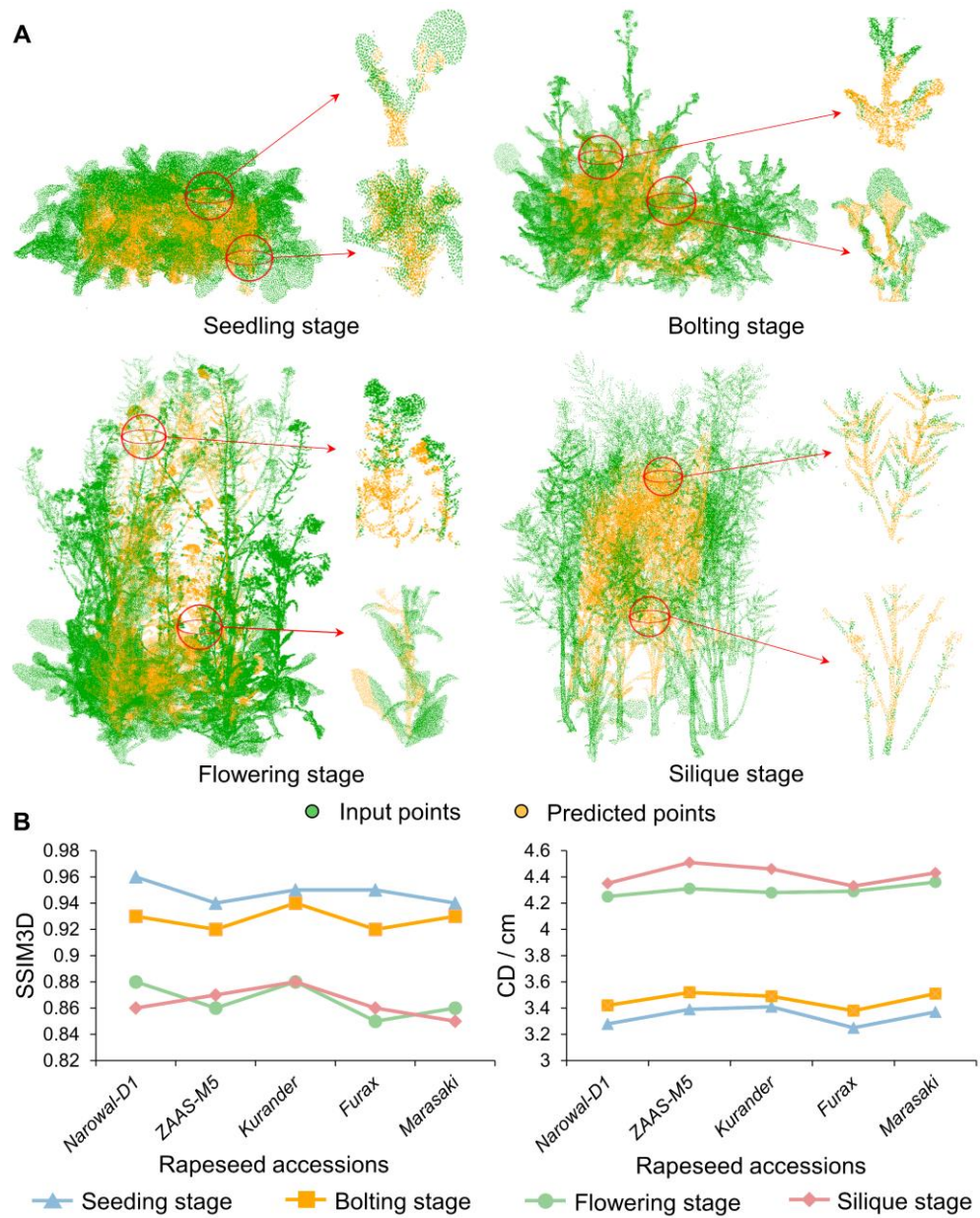

**Figure 4.** Performance of CP-PCN on field-grown rapeseed populations. (**A**) Point cloud completion for field-grown rapeseed plot: the incomplete point cloud (green) is completed by CP-PCN (orange), with detailed views of leaf, flower, silique and stem architectures. (B) Quantitative evaluation of CP-PCN across five rapeseed accessions, showing the chamfer distance (CD) and three-dimensional structural similarity index (SSIM3D) values.

For a more comprehensive evaluation, CP-PCN was applied to five different rapeseed accessions selected based on variations in plant height and branching. The quantitative evaluation results, presented in Figure 4B, show that CP-PCN achieved consistent performance across all accessions in terms of both chamfer distance (CD) and three-dimensional structural similarity index (SSIM3D) for the four growth stages. The model demonstrated high accuracy in predicting the canopy architecture, with the average CD and SSIM3D values for all accessions showing only slight variations across the growth stages. Specifically, at the seedling stage, the average CD was 3.34 cm and the SSIM3D value was 0.95; at the bolting stage, the average CD was 3.46 cm, with an SSIM3D value of 0.93; at the flowering stage, the average CD was 4.30 cm, and SSIM3D was 0.87; and at the silique stage, the average CD was 4.42 cm, with an SSIM3D of 0.89. Notably, the model maintained good performance across all accessions at the seedling, bolting, flowering, and silique stages, indicating its ability to handle different canopy architectures effectively. However, as the canopy architecture became more complex in the later stages (flowering and silique), a slight decrease in both CD and SSIM3D was observed, highlighting the challenges associated with reconstructing more intricate canopy architectures. Despite this, the model's performance remained robust, confirming its potential for accurate canopy reconstruction across a variety of growth stages and plant architectures.

**Comparison with other models**

Figure 5A presents the comparison results of rapeseed point cloud completion across different models. The CP-PCN consistently provided the highest accuracy, capturing fine canopy and silique features with high fidelity, and producing 3D reconstructions that closely matched the ground truth. In contrast, PF-Net and GRNet struggled to reconstruct detailed branch architectures and siliques, with considerable discrepancies between the predicted and ground truth point clouds. Although the PoinTr model achieved relatively better results in some stages, it faced difficulties in complex canopy formations, often missing critical architectural elements.

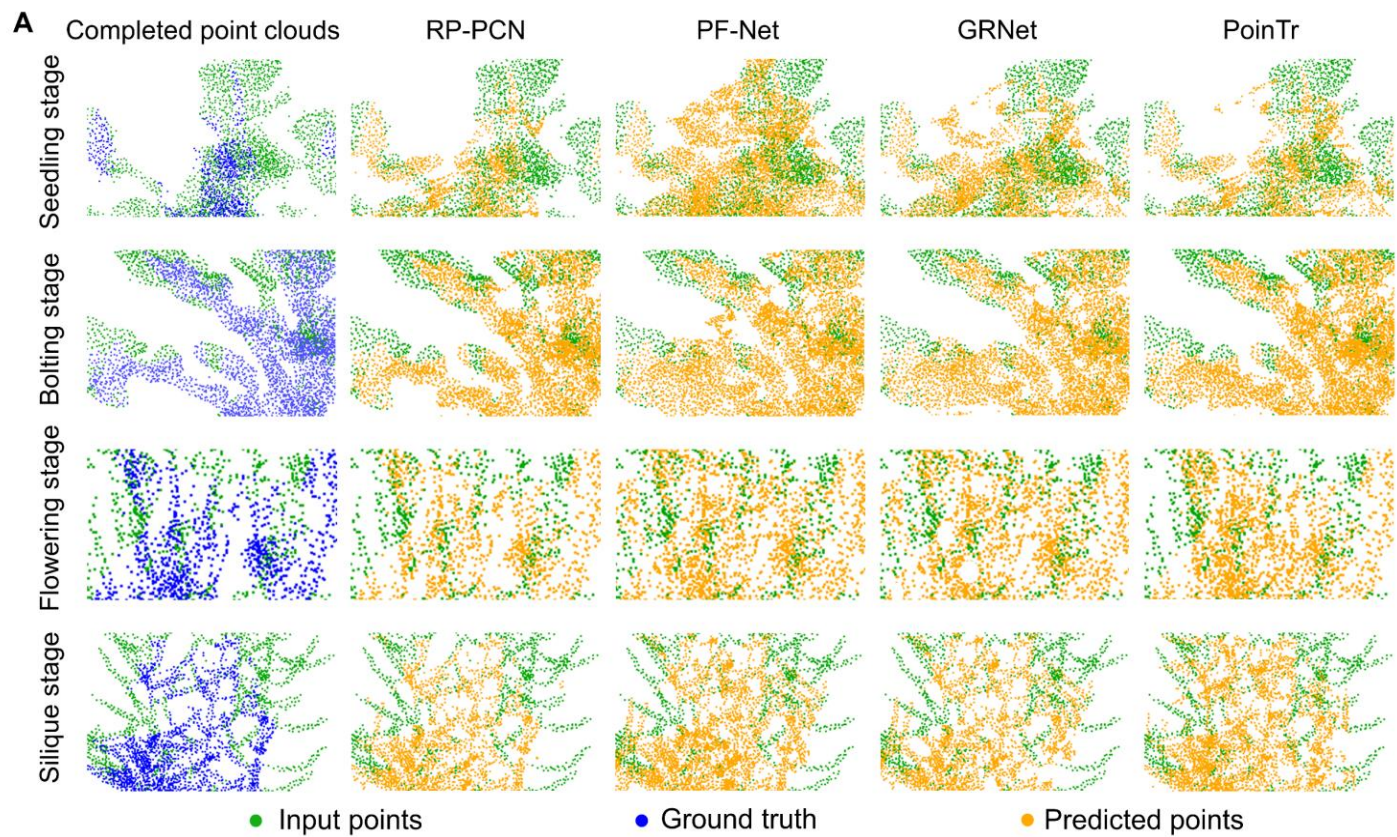

**Figure 5.** Visualization and comparison of rapeseed population point cloud completion models. (**A**) Visualization of completed point clouds at four growth stages (seedling, bolting, flowering, and silique) for CP-PCN, PF-Net, GRNet, and PoinTr. The input points, ground truth, and predicted points are represented as green, blue, and orange dots, respectively. (**B**) Quantitative evaluation of the Chamfer distance (CD) values for CP-PCN, PF-Net, GRNet, and PoinTr across different growth stages. (**C**) Computational comparison of the models in terms of CD and floating-point operations per second (FLOPs), with model size indicated.

Quantitative evaluations further confirmed these observations (Figure 5B). The CP-PCN model outperformed all other models across all stages, with CD values of 3.35 cm, 3.46 cm, 4.32 cm, and 4.51 cm for the seedling, bolting, flowering, and silique stages, respectively, resulting in an overall average CD of 4.05 cm. In contrast, PF-Net and GRNet exhibited high CD values, indicating weaker reconstruction abilities, particularly in later growth stages. During the silique stage, CD values for PF-Net and GRNet were 11.98 cm and 8.34 cm, respectively. Although PoinTr achieved improved accuracy, it still yielded higher CD values than CP-PCN.

In addition to accuracy, the computational efficiency and model size of each algorithm were evaluated (Figure 5C). CP-PCN effectively balanced high accuracy with computational efficiency, making it a suitable choice for large-scale, high-throughput field applications. In comparison, PF-Net was computationally efficient due to its simplicity, but its poor performance in point cloud completion limited its application in high-precision tasks. PoinTr and GRNet, while achieving better accuracy, incurred substantial computational costs and had large model sizes. Specifically, PoinTr required about 420 giga floating point operations per second (GFLOPs), leading to longer inference times.

**Estimation of rapeseed yield**

The SEI values derived from complete population point clouds exhibited a higher correlation with rapeseed yield compared to those derived from incomplete point clouds as shown in Figure 6. Regression analysis revealed that SEI calculated from the complete point cloud data achieved the $R^2$ of 0.90, significantly improving the yield estimation accuracy, compared to 0.81 for incomplete point cloud data. This improvement highlights the importance of the CP-PCN model in enhancing the precision of yield prediction. In addition, the SEI derived from the middle silique zone demonstrated the highest correlation with yield, with $R^2$ values increasing from 0.69 to 0.79, further underscoring the crucial role of complete point clouds in refining yield estimation. This finding is consistent with previous research, which indicates that siliques from the middle canopy layers contribute most significantly to yield (Lin et al., 2024). In contrast, the SEI calculated from the lower and upper silique regions showed lower $R^2$ values, indicating their relatively minor contribution to the overall yield, possibly influenced by environmental and genetic factors.

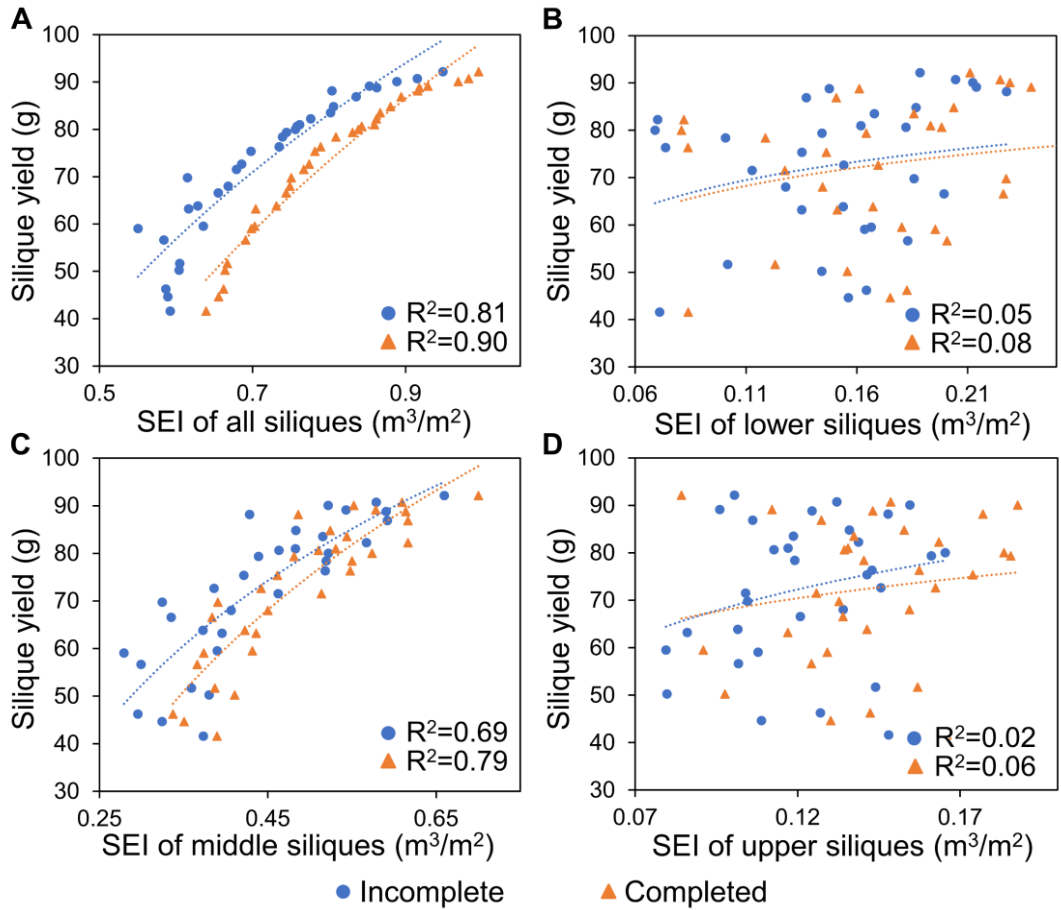

**Figure 6.** Regression analysis between silique efficiency index (SEI) and silique yield based on complete and incomplete point clouds. (**A**) SEI derived from all silique layers. (**B**) SEI from the lower layer. (**C**) SEI from the middle layer. (**D**) SEI from the upper layer.

## Discussion

**Importance of the high-quality annotated data for modeling**

The quality of the training dataset is a critical factor for the performance of deep learning models. A high-quality dataset must be representative, encompassing a broad spectrum of target scenarios, sufficiently diverse to reflect the range of states and conditions present in the target, and must feature precise annotations to prevent the model from learning biased or erroneous patterns (Paullada et al., 2021). In constructing the rapeseed population point cloud completion dataset, we firstly proposed a novel simulation method, referred to as VRI, which produced more realistic simulation results by extracting dynamic nature of rapeseed canopy architectures throughout growth compared to the conventional approach of replicating individual plant data. During the seedling stage, the canopy architectures of different individuals within the same accession are relatively uniform, making traditional replication methods reasonably effective (Sun et al., 2021). However, as plants enter the flowering and silique stages, even individuals of the same accession exhibit substantial variability in canopy architecture, and simple replication of individual plants fails to capture this true diversity at the population level. Moreover, the proposed occlusion point detection algorithm successfully identified actual occlusion points from the simulated complete canopy point clouds, which should be included for model learning in the training dataset. This approach follows the principles of multi-view 3D reconstruction, where UAV-based imaging typically captures only surface information, while internal architectures remain occluded and unreconstructed. Existing methods often create datasets by randomly clipping sections from complete point clouds, which results in many surface points being incorrectly included as targets for completion and finally reducing overall accuracy (Chen et al., 2023, Huang et al., 2020). By focusing on occluded points, our method ensures the training dataset better represents real-world conditions, improving model performance.

**Influence of hyper-parameters, DGCFE and MRDG on CP-PCN**

The results from the ablation study highlight the crucial roles that both the DGCFE and MRDG modules play in enhancing point cloud completion performance in CP-PCN (Supplemental Figure 1). These findings align with the dynamic and complex nature of rapeseed canopy growth, where structural variations arise at different growth stages. During the early stages of growth, rapeseed canopies exhibit relatively simple geometric features, which require effective local feature extraction for accurate modeling. The DGCFE module, which specializes in this task, significantly enhances the model's ability to recover fine details, particularly in sparse or irregular canopy architectures. As growth progresses, the canopy's structural complexity increases, necessitating the MRDG module to model hierarchical relationships and capture broader architectural features that emerge during later stages. The CP-PCN's ability to dynamically adjust to these varying complexities distinguishes it from other methods. Without the DGCFE and MRDG components, model performance drops significantly, as evidenced by the elevated CD values of 6.04 cm and 6.21 cm during the flowering and silique stages, respectively. This performance degradation emphasizes the importance of these modules in ensuring robust canopy reconstruction across all growth stages. However, despite the inclusion of DGCFE and MRDG, the model's performance diminishes as the canopy architecture becomes more complex and internal occlusions intensify. This results in a gradual decline in point cloud completion accuracy, suggesting that the current model's learning capability is not infinite. The model still struggles to extract relevant feature representations, particularly in regions with severe occlusions, where the canopy's density and overlap of leaves pose additional challenges. These results highlight CP-PCN's modular design, which effectively combines local feature extraction and multi-resolution strategies to address the challenges posed by complex plant canopy architectures. Nonetheless, further refinements in occlusion handling and feature extraction in densely occluded regions will be crucial for improving the model's generalization and accuracy, especially under real-world conditions with significant canopy overlap.

In addition to the ablation study, a sensitivity analysis of key hyper-parameters, including the number of nearest neighbors ($k$), the number of encoder layers ($O$), and the input and output point cloud sizes ($N$ and $M$), provided valuable insights into CP-PCN's performance (Supplemental Figure 2). The $k$ was found to stabilize the model's performance when set above 20, indicating that this value was optimal for capturing spatial relationships without unnecessary computational overhead. Similarly, the $O$ was most effective at five layers, providing an ideal balance between depth and computational efficiency. Fewer layers limited the model's ability to extract high-level features, while deeper layers led to excessive complexity without additional accuracy. Furthermore, the analysis revealed that increasing the $N$ and $M$ beyond 8192 did not result in significant improvements in model accuracy. Larger point clouds only increased point density, which did not enhance the model's ability to capture structural details. These insights suggest that CP-PCN operates efficiently with moderate point cloud sizes, confirming that optimal hyper-parameter selection contributes to both model accuracy and computational efficiency.

**Contribution of CP-PCN on crop population canopy analysis**

The CP-PCN model proposed in this study introduces a novel and highly effective approach for obtaining complete architectural information of plant population canopies. Experimental validation with rapeseed canopies demonstrated that point clouds completed by CP-PCN significantly outperformed incomplete point clouds in terms of yield estimation accuracy. Traditionally, rapeseed yield has been estimated using silique-related traits, such as number, length, and weight, which are typically measured without considering the spatial distribution of siliques within the canopy (Zheng et al., 2022). However, the spatial heterogeneity of the canopy and the unique plant architecture of rapeseed at different growth stages make this estimation challenging. The complete point cloud of rapeseed population by using CP-PCN enables a more accurate representation of the rapeseed canopy. In our study, we divided the canopy of rapeseed into three vertical layers, including upper, middle, and lower layers, to evaluate the yield contribution of each layer. The results showed that siliques in the middle layer had the strongest correlation with overall yield, indicated by their higher silique volume and more efficient light capture compared to the upper and lower layers. Quantification of canopy architectures would help on machine-based description of canopy formation and development, which would open new doors of agronomic trait collection and yield estimation criteria. The parameters extracted by CP-PCN would be used as new agronomic digital traits and further co-analyzed with genomic information for the identification of novel genetic loci. By using digital traits as the input, high-throughput and time-series identification of genetic trait transition switches could be likely uncovered, which would shed light on creative ways of genetic improvement for ideal canopy architectures under different circumstances.

    The code and dataset preparation methods used for model training in this study have been fully made available for public access. To apply the CP-PCN model to other crops, users only need to follow the procedures outlined in this research to recreate the necessary dataset and train the model accordingly. The time required for this step varies based on the sample size and available hardware resources, and can span a wide range. As a reference, Supplemental Note 2 provides the time required for dataset preparation and model training under the hardware conditions used in this study. Once the model is trained, it can be directly applied to new crops without the need for further adjustments. In terms of model application, the time required to complete a single rapeseed field plot is approximately 350 seconds, with most of this time spent on pose

estimation and 3D reconstruction. The actual point cloud completion, however, only takes a few seconds using the CP-PCN. While the overall processing time per plot is several minutes, this duration is acceptable for high-throughput phenotyping applications where real-time processing is not typically a strict requirement. If lidar-based devices were used to directly capture the crop surface point clouds, the need for 3D reconstruction would be eliminated, allowing the raw point clouds to be input directly into the model. This would reduce the time required for point cloud completion to just the inference time, which takes only a few seconds.

## Materials and methods

### Experimental design and data acquisition

Two experimental setups were designed for data acquisition: a field experiment and a greenhouse potted plant experiment. The field experiment aimed to capture 3D models of individual plants and surface point clouds of rapeseed populations under field conditions. The potted plant experiment assessed the differences in canopy architecture between plants grown in controlled and field environments. The field experiments were conducted at the Jiaxing Academy of Agricultural Sciences, Zhejiang Province, China (120°41′39″E, 30°51′14″N), with 300 rapeseed accessions planted in three replicates, resulting in 900 experimental plots (Supplemental Figure 3). Each plot contained 16 plants arranged with 28 cm row spacing and 25 cm plant spacing. Uniform growth conditions were maintained with optimized water and nutrient management. Additionally, four rapeseed accessions with distinct architectures were randomly selected for potted cultivation at the Zhejiang University Agricultural Experimental Station, Hangzhou, China (120°4′46″E, 30°18′31″N).

For field-grown individual plants, one plant was randomly selected from the center of each of 100 plots (chosen from one replicate of the 300 accessions) at four growth stages: seedling, bolting, flowering, and silique. This yielded a total of 400 potted plants (100 per stage). UAV imaging was performed using a DJI Mavic 3T UAV (DJI, Shenzhen, China) to capture high-resolution videos at 30 frames per second (3840 × 2160 pixels). The UAV circled each potted plant three times at 0°, 30°, and 60° angles to ensure full surface coverage, confirmed by preliminary tests to generate accurate 3D models. The same approach was used to capture the 3D models of the potted plants.

For field-level rapeseed populations, a UAV with a -60° viewing angle and a 5-meter distance from the plot center was used for multi-view imaging. Images were captured every 10° along the flight path, resulting in 36 images per plot (8000 × 6000 pixels). Thirty-two plots were randomly selected from the 300 non-destructive sampling plots for testing the similarity between simulated and real-world surface point clouds, as well as for yield prediction based on silique stage data. Additionally, five plots were selected for destructive sampling. After UAV imaging, the spatial relative positions of all plants within each plot were recorded. The plants were then excavated from the field plots, transplanted into pots, and individually scanned using UAV video for further testing of the CP-PCN model's ability to complete point clouds for field-grown rapeseed populations.

**Dataset generation for rapeseed population point cloud completion model**

To develop the rapeseed population point cloud completion model, we proposed virtual-real integration (VRI) simulation method combined with an occlusion point detection algorithm to generate a comprehensive training dataset. Multi-view videos of individual rapeseed plants at four key growth stages (seedling, bolting, flowering, and silique) were captured using a UAV. Images were sampled at regular intervals, ensuring full coverage of the plant surfaces from different perspectives (Figure 1A). The structure from motion (SfM) algorithm was used to calculate the pose information for each image, which, along with the images, were input into the Instant-NGP model, a neural radiance field (NeRF) approach, to generate 3D models of the individual plants, including both point clouds and triangular mesh models. (Jiang et al., 2020).

The VRI method simulated the real field plot configuration with 16 plants spaced 25 cm apart in rows of 28 cm. Plants from the individual dataset at the same growth stage were randomly selected, and the base of each plant's bounding box, where the main stem intersects the root, was aligned with the simulated plot coordinates. This method generated diverse rapeseed population canopy architectures (Figure 1B). To create a point cloud completion dataset, we developed an occlusion point detection algorithm based on ray tracing. This algorithm identified surface points (visible from at least one camera perspective) and occluded points (completely invisible due to inter-plant occlusion). The process calculates ray intersections between each point in the plant cloud and camera positions, determined by simulating flight altitude and imaging angles. The surface points were used as input data, and occluded points were the model's target output for completion (Figure 1C). Using this methodology, 4,000 training samples were generated, and the dataset was split into 80% for training and 20% for validation.

**Algorithm 1** occlusion point detection

**Input**: point cloud and triangle mesh of plant canopy, imaging coordinates.

**Output**: surface points, occluded points.

1:   for each point in the plant point cloud:
2:       for each imaging point:
3:           for each triangle in the mesh model:
4:               if a ray from the plant point to the imaging point intersects with the triangle:
5:                   Increment intersection count by 1;
6:               end if
7:           end for
8:           if intersection count > 0:
9:               Increment occluded direction by 1;
10:          end if
11:      End for
12:      if occluded direction > total imaging points:

| 13: | Mark the point as completely occluded and unavailable from multi-view 3D reconstruction; |
| 14: | end if |
| 15: | end for |

For five destructively sampled field plots, the individual plant videos were processed using the same 3D reconstruction pipeline to generate complete canopy models. These models were assembled using the VRI method, with the recorded relative positions of each plant, and the ground truth of the complete point cloud for these plots was obtained.

**Evaluation of similarity between simulated and real point clouds**

The three-dimensional structural similarity index (SSIM3D) was utilized to assess the similarity between the surface point clouds from UAV multi-view images and simulated population point clouds (Alexiou and Ebrahimi, 2020). The calculation formula is as follows:

$$SSIM3D(X, Y) = \frac{1}{N} \sum_{P=1}^{N} \frac{|F_X(q) - F_Y(p)|}{\max(|F_X(q)|, |F_Y(p)|) + \varepsilon} \quad (1)$$

where $X$ and $Y$ represent two sets of 3D point clouds, $F_X(q)$ and $F_Y(p)$ denote the feature values of points $q$ and $p$, respectively, and $N$ is the number of points in the point cloud. The term $\varepsilon$ is a small value to prevent division by zero.

For each of the 32 real field plots, SSIM3D values were calculated by comparing the UAV-derived surface point clouds with 1,000 simulated plots. The highest similarity score from these comparisons was retained as the final evaluation metric (Supplemental Figure 4). To ensure meaningful comparison, simulated point clouds were first processed with the occlusion point detection algorithm, and only the visible surface point clouds were included in the comparison. This process ensured that the comparison focused solely on the observable parts of the canopy, thereby reflecting the accuracy of the simulated canopy architecture in relation to real-world field data. The overall similarity score for each simulation approach was computed by averaging the highest SSIM3D values across all 32 plots. In addition to the VRI simulation, two conventional methods based on individual plant replication were also evaluated: one using potted plants and the other using field-grown plants. Given the morphological differences between these plant types and the variations across different growth stages, comparisons were made separately for each stage, enabling a detailed evaluation of each method's performance in replicating realistic canopy architectures at different developmental stages.

**Network architecture**

To address the highly variable and complex structural characteristics of rapeseed population canopy architectures across different growth stages, we designed a deep learning network called CP-PCN, which integrates the advantages of GANs and graph neural networks (Figure 7B). The CP-PCN architecture consists of two primary components: the generator and the discriminator. The generator predicted the missing portions of the point cloud based on the incomplete input, while the discriminator evaluated the plausibility of the generated point cloud by distinguishing it from the ground truth data. To preserve plant architecture during down-sampling, a sliding window approach was employed to segment large population point clouds into smaller, manageable blocks before processing (Figure 7A). This strategy ensured structural integrity and enabled more effective learning of local and global canopy features.

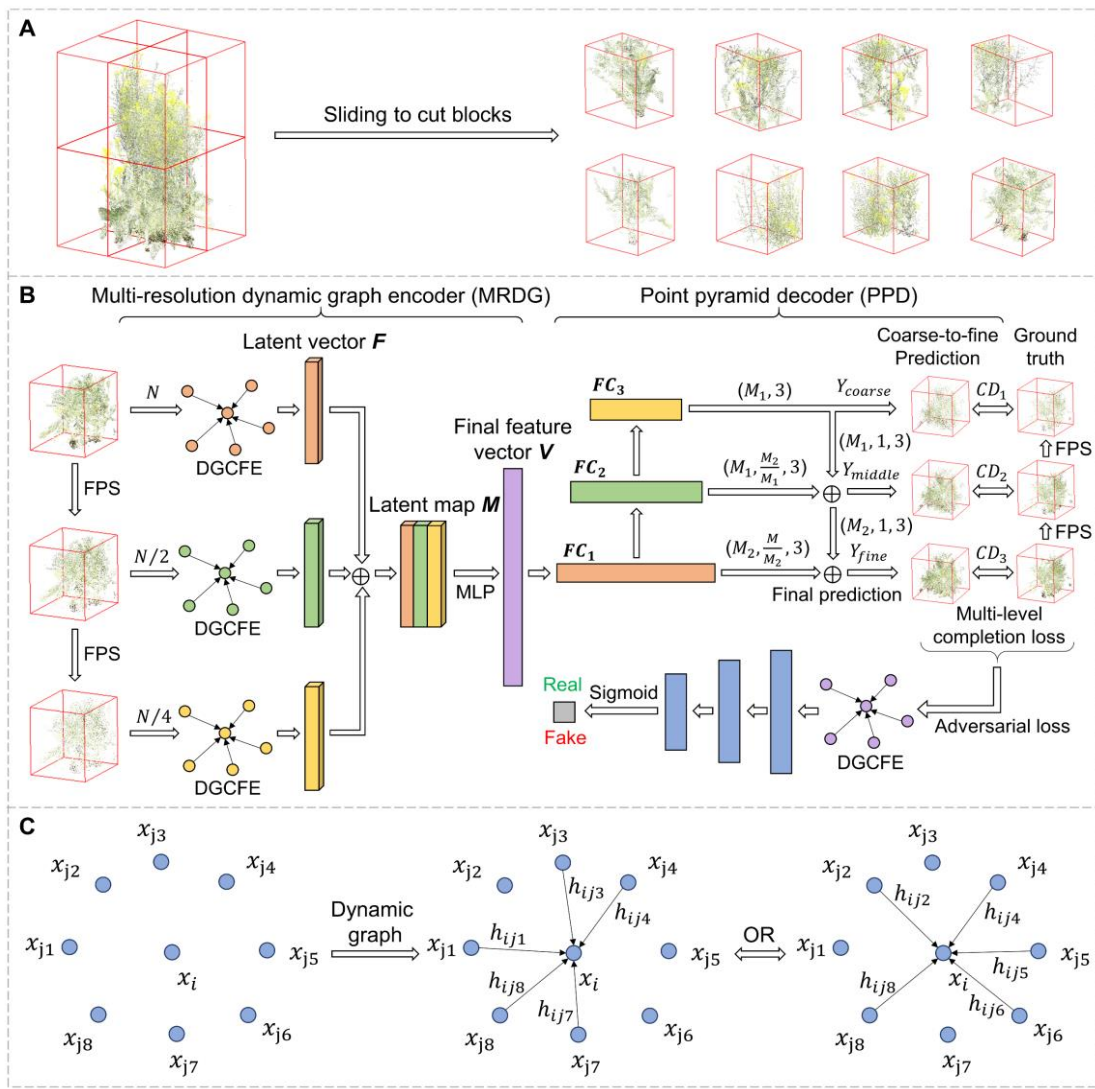

**Figure 7.** Workflow of the crop population point cloud completion network (CP-PCN). (**A**) The input rapeseed population point cloud is evenly divided into eight blocks using a sliding window strategy. (**B**) The CP-PCN adopts a generative adversarial network (GAN) framework, where the generator consists

of a multi-resolution dynamic graph encoder (MRDG) and a point pyramid decoder (PPD). Before encoding, the input point cloud is down-sampled into three resolutions using iterative farthest point sampling (FPS). Feature extraction at each resolution is performed by the dynamic graph convolutional feature extractor (DGCFE). (**C**) DGCFE can dynamically adjust the feature relationship between the center point and adjacent points.

The input to the network is a point cloud with a size of $N = 81928$ points, which is first down-sampled twice using iterative farthest point sampling (IFPS) (Charles et al., 2017). This down-sampling results in two additional resolutions, $N/2 = 4096$ and $N/4 = 2048$, capturing both fine and global canopy features at varying scales. After down-sampling, multi-resolution feature extraction is performed by a multi-resolution dynamic graph encoder (MRDG). The MRDG utilizes a dynamic graph convolutional feature extractor (DGCFE) to learn both local and global spatial relationships of the points across the three resolutions (Wang et al., 2019). This enables the model to adaptively update the neighborhood relationships and capture fine-grained geometric features at different scales (Figure 7C). The resulting feature vectors from each resolution are then concatenated to form a 5760-dimensional latent vector $V$. This feature vector serves as a rich representation of the canopy architecture, enabling the network to effectively handle the complexity and variability of plant architectures at various growth stages.

The point pyramid decoder (PPD) is then employed to decode this feature vector into three separate point clouds at three different resolutions: coarse, middle, and fine. The coarse point cloud $Y_{coarse}$ contains the primary center points, while the middle and fine point clouds $Y_{middle}$ and $Y_{fine}$ represent more detailed, refined architectures. The predicted point clouds at these resolutions have sizes of 2048, 4096, and 8192 points, respectively. These point clouds are progressively refined and up-sampled using fully connected layers in the PPD, ensuring that the final output closely resembles the ground truth, particularly in areas where canopy details were occluded. Finally, the completed point cloud is generated by merging the input point cloud with the predicted point cloud at the fine resolution. For more detailed information about the network architecture, including the MRDG and PPD, please refer to Supplemental Note 3.

**Network training, validation and testing**

To train the CP-PCN model for point cloud completion, an effective loss function was designed to optimize both the feature extraction and generation processes. The total loss function consisted of two components: a multi-stage completion loss and an adversarial loss. The completion loss was based on the chamfer distance (CD), a widely used metric for point cloud similarity, and was computed between the predicted point clouds and the ground truth at three hierarchical resolutions (Wu et al., 2021). The formula for CD is:

$$d_{CD}(S_1, S_2) = \frac{1}{S_1} \sum_{x \in S_1} \min_{y \in S_2} \|x - y\|_2^2 + \frac{1}{S_2} \sum_{y \in S_2} \min_{x \in S_1} \|y - x\|_2^2 \qquad (2)$$

where $S_1$ and $S_2$ represented two sets of 3D point clouds. Accordingly, the

multi-stage completion loss was defined as:

$$L_{com} = d_{CD1}(Y_{fine}, Y_{GT}) + \alpha\, d_{CD2}(Y_{middle}, Y'_{GT}) + 2\alpha\, d_{CD3}(Y_{coarse}, Y''_{GT}) \qquad (3)$$

where $Y_{fine}$, $Y_{middle}$, $Y_{coarse}$ are the predicted point clouds at three resolutions, and $Y_{GT}$, $Y'_{GT}$, $Y''_{GT}$ are the corresponding ground truths obtained by sub-sampling using IFPS. The hyperparameter $\alpha$ was adjusted during training to balance contributions from different resolutions.

To further enhance the realism of the predicted architectures, an adversarial loss based on a GAN framework was incorporated. The generator $F()$ aimed to reconstruct the missing region, while the discriminator $D()$ attempted to distinguish the generated output from real samples. The adversarial loss was formulated as:

$$L_{adv} = \sum_{1 \leq i \leq S} \log(D(y_i)) + \sum_{1 \leq i \leq S} \log\left(1 - D(F(x_i))\right) \qquad (4)$$

where $x_i$ and $y_i$ represent the input and ground truth samples respectively, and $S$ is the batch size. The total loss function for model training was:

$$L = \lambda_{com} L_{com} + \lambda_{adv} L_{adv} \qquad (5)$$

with $\lambda_{com} = 0.9$ and $\lambda_{adv} = 0.1$, ensuring a stronger emphasis on structural accuracy during training. For a more detailed loss function design process, please refer to Supplemental Note 4.

The CP-PCN network was implemented using PyTorch and trained on a workstation equipped with an Intel i9-10900K CPU and an Nvidia RTX 3090 GPU. The Adam optimizer was used with an initial learning rate of 0.0001 and a batch size of 8. The model was trained on the simulated rapeseed population dataset, consisting of annotated surface and occluded point clouds. All network weights were randomly initialized due to the absence of publicly available pre-trained models for this task. Training was conducted for up to 200 epochs or until the loss dropped below 0.1.

During training, batch normalization and ReLU activation were used in the MRDG and discriminator, while the PPD utilized only ReLU activations except at the output layer. The weighting parameter $\alpha$ for the multi-resolution loss was dynamically adjusted: 0.01 for epochs 0–30, 0.05 for epochs 30–80, and 0.1 for epochs beyond 80. The validation set was evaluated every 10 epochs; if the loss decreased, the model parameters were updated to prevent overfitting and ensure generalization.

Once trained, the CP-PCN model was applied to the surface point clouds of five destructively sampled rapeseed plots reconstructed from UAV imagery. The model completed the occluded regions, and the predicted points were merged with the input to form complete 3D reconstructions. These predictions were then quantitatively compared with the ground truth to evaluate the model's real-field performance (Supplemental Figure 5).

**Ablation study and hyper-parameter analysis**

To evaluate the contribution of each key component in the proposed CP-PCN architecture, a series of ablation studies were conducted. First, to examine the effectiveness of the multi-resolution encoding strategy, we replaced the MRDG with a single-resolution encoder. Second, to evaluate the impact of the

DGCFE, we substituted it with a standard MLP. Both ablation variants were trained and tested under the same conditions as the original model, and their performance was compared in terms of point cloud completion accuracy.

In addition, we conducted a sensitivity analysis on four hyper-parameters: (i) the number of nearest neighbors ($k$) in dynamic graph convolution, tested with values $\{8, 10, 12, 16, 20, 24, 28, 32\}$; (ii) the input ($N$) and output ($M$) point cloud sizes, varied across $\{2048, 4096, 8192, 16384\}$; and (iii) the number of layers ($O$) in the MRDG, set to $\{4, 5, 6, 7\}$. Based on prior research, the first five layers were set to $\{64, 128, 256, 512, 1024\}$. For deeper configurations with six and seven layers, the additional layers were set to 2048 and 4096, respectively, to further enhance feature extraction capacity. Each parameter was varied independently while keeping the others constant to isolate its effect on model performance.

**Comparison analysis**

To further evaluate the effectiveness of CP-PCN, we benchmarked it against several state-of-the-art point cloud completion models, including PF-Net, GRNet, and PoinTr (Xie et al., 2020; Huang et al., 2020; Yu et al., 2021). These models were selected as they represent distinct paradigms in point cloud completion. PF-Net, a lightweight model, utilizes a GAN-based architecture with MLP-based feature extraction to generate missing regions. GRNet employs a deep residual network structure, inspired by ResNet, to enhance precision through hierarchical feature learning. In contrast, PoinTr incorporates a transformer-based architecture, achieving high accuracy but at the cost of increased computational complexity. All models were trained and tested under identical conditions, ensuring a fair comparison. Performance was evaluated based on visual inspection, CD, and computational efficiency, allowing for a comprehensive assessment of reconstruction quality and model efficiency.

**CP-PCN for yield estimation**

To validate the agronomic utility of the CP-PCN model, we evaluated whether the completion of occluded canopy architectures could improve the accuracy of yield estimation in field-grown rapeseed populations. Given that silique architecture is a major determinant of yield, we proposed a new index termed the SEI, defined as the silique volume per unit ground area of the plot (Figure 1E). The silique volume was extracted from the completed population point clouds using a previously developed segmentation model (PST) and computed via voxel-based volume estimation (Du et al., 2023). The SEI was calculated as follows:

$$Silique\ efficiency\ index = silique\ volume\ /\ plane\ area \qquad (6)$$

To assess the correlation between SEI and actual yield, 32 non-destructively sampled plots were selected. At full maturity, siliques from these plots were manually harvested, dried, and weighed to obtain ground-truth yield values. The SEI was computed for both incomplete and completed point clouds, and regression analysis was conducted to evaluate their respective prediction accuracies. The coefficient of determination $R^2$, used to quantify the relationship between predicted and observed yield.

## Acknowledgements

This work was funded by the National Natural Science Foundation of China (32371985) and the International S&T Cooperation Program of China (2024YFE0115000). We extend our heartfelt gratitude to Ruisen Wang, Yi Feng, Mengjie Gong and Guangyu Wu, for their participation in the experiments, and to the Jiaxing Academy of Agricultural Sciences for their assistance with the experimental data acquisition.

## Availability of supporting data and source code

All source codes and test data involved in this study are available on GitHub (https://github.com/Ziyue-Guo/RP-PCN.git).

## Declaration of Competing Interest

The authors declare that they have no known competing financial interests or personal relationships that could have appeared to influence the work reported in this paper.

## Contributions

Z. G. designed the study, conducted the experiments, and wrote the manuscript. Y. S. contributed to the experimental design and data collection. X. Y. assisted with model development and data analysis. Y. Z. and L. J. provided the experimental materials and contributed to the experimental design. H. C. supervised the study and revised the manuscript.

## Supplemental Figure

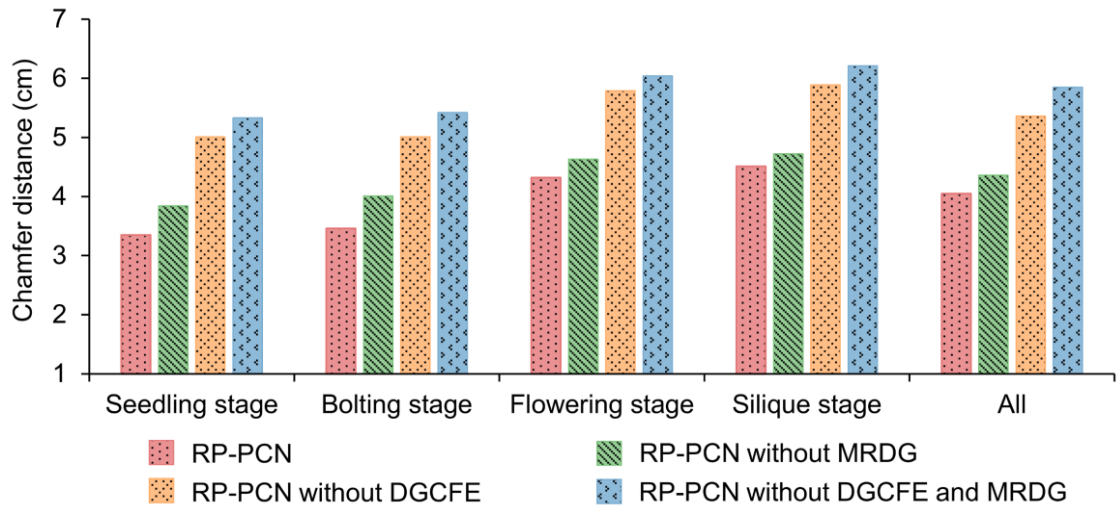

**Supplemental Figure 1.** Results of the ablation study on crop population point cloud completion network (CP-PCN) with different architectures, trained with datasets from seedling, bolting, flowering, and silique stages, as well as all stages. The chart compares the performance of the full CP-PCN model with versions excluding the dynamic graph convolutional feature extractor (DGCFE) and the multi-resolution dynamic graph (MRDG) individually, as well as a version excluding both components.

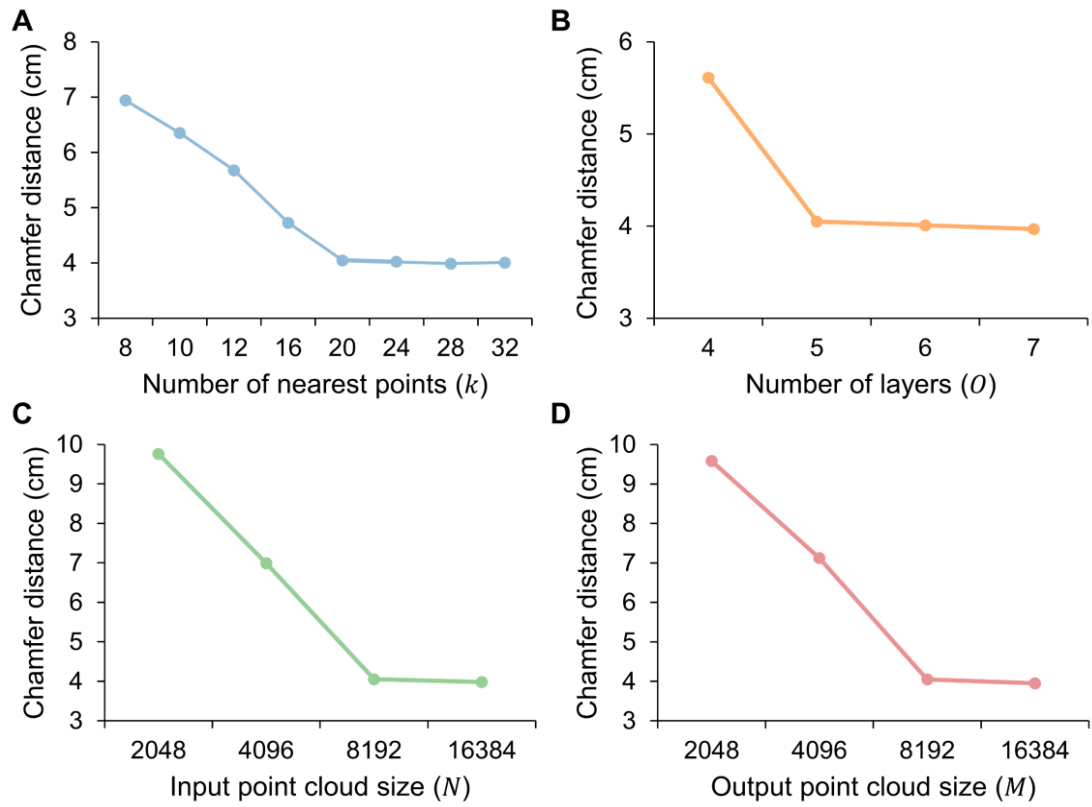

**Supplemental Figure 2.** Sensitivity analysis of (a) the number of nearest neighbors ($k$), (b) number of layers in the encoder ($O$), (c) input point cloud size ($N$), and (d) output point cloud size ($M$) on the model performance.

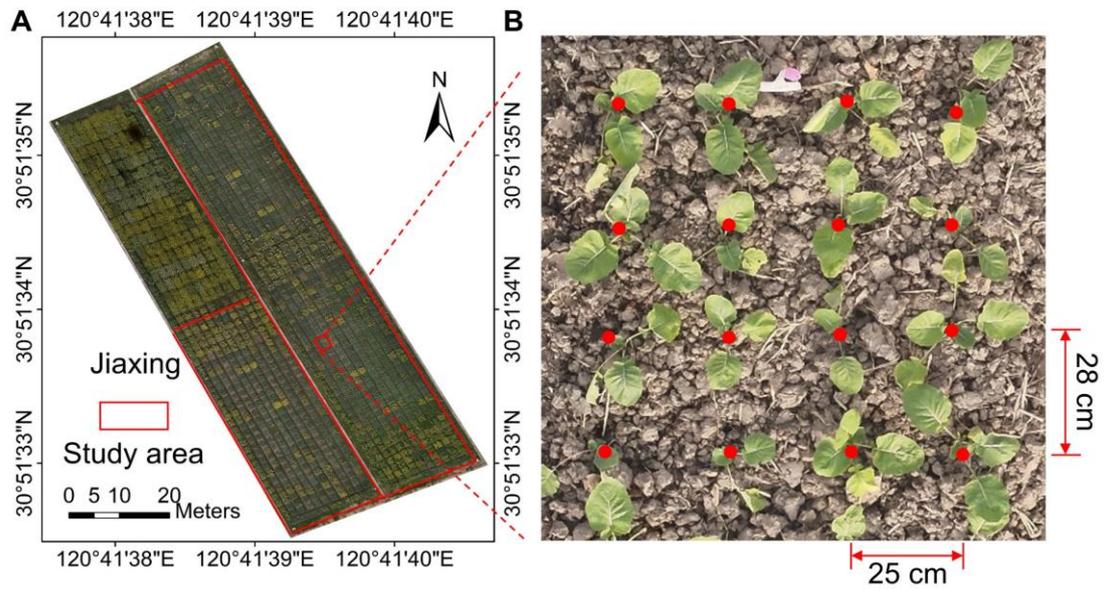

**Supplemental Figure 3.** Location of the experimental field (**A**) in Jiaxing, Zhejiang province, China with (**B**) the rapeseed plant layout within a plot. Each plot contains 16 rapeseed plants with the spacing of 28×25 cm.

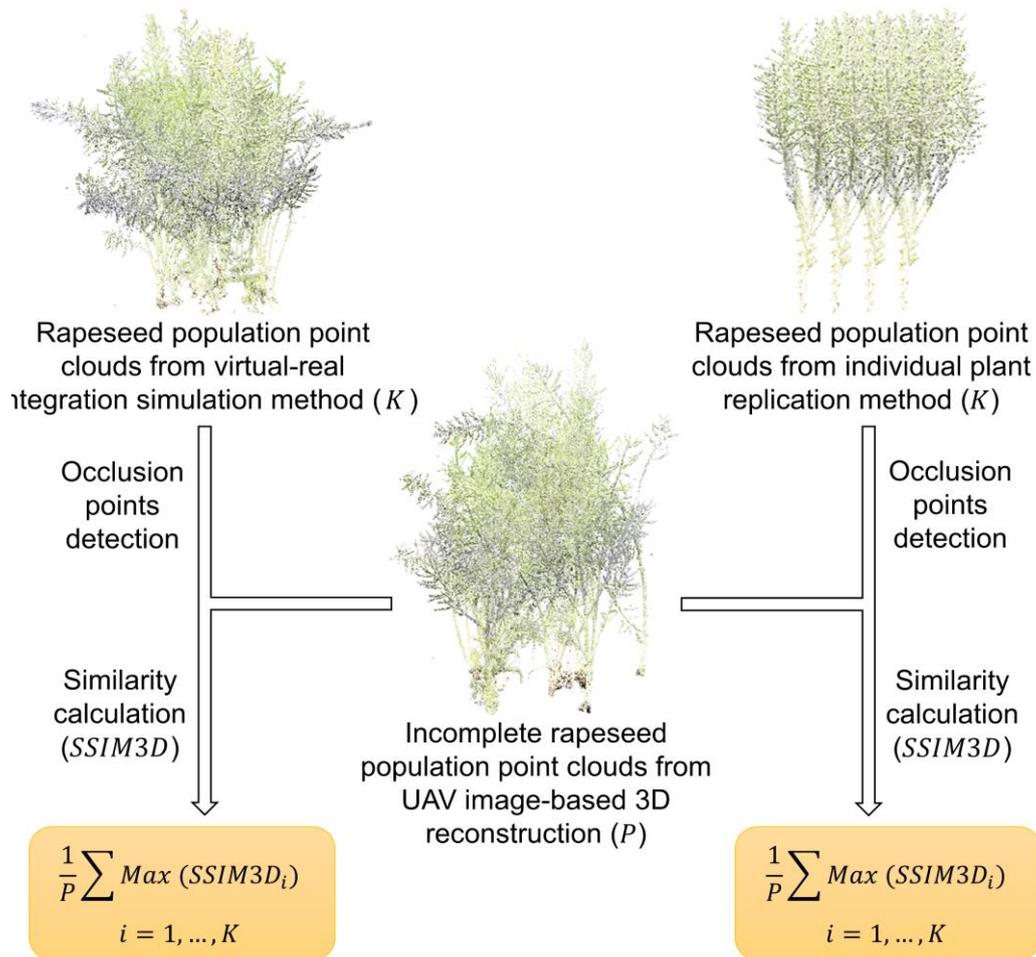

**Supplemental Figure 4.** Quantitative evaluation of simulated rapeseed population point clouds by comparing them with UAV-derived ground-truth using the three-dimensional structural similarity index (SSIM3D). The proposed virtual-real integration (VRI) simulation method was compared against traditional individual plant replication approaches.

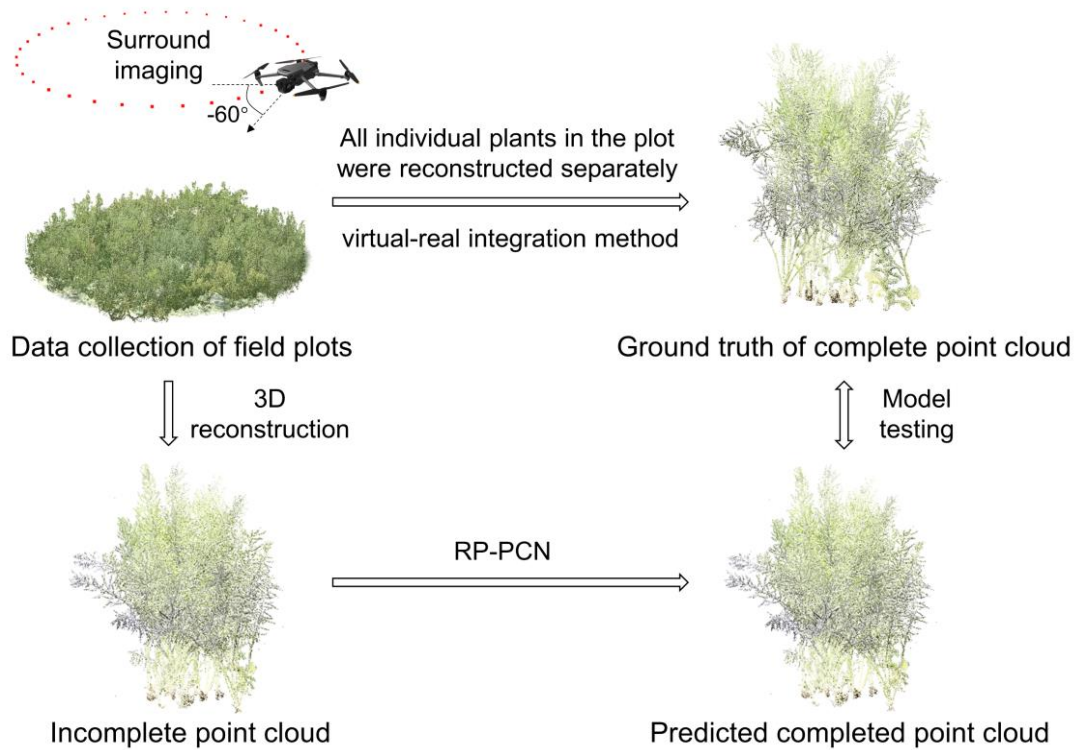

**Supplemental Figure 5.** Model testing pipeline of the crop population point cloud completion network (CP-PCN) using field plot data.

# Supplemental Note

**Supplemental Note 1. Visualization of CP-PCN completion results**

Figure N1 presents the point cloud completion results generated by CP-PCN at different growth stages. Through slicing along the X, Y, and Z planes, the internal completion of the crop population canopy is showcased. It is evident that CP-PCN performs well in reconstructing both relatively simple architectures during the seedling stage and the more complex architectures during the flowering and silique stages. This highlights the model's robust capability to effectively handle a wide range of canopy complexities.

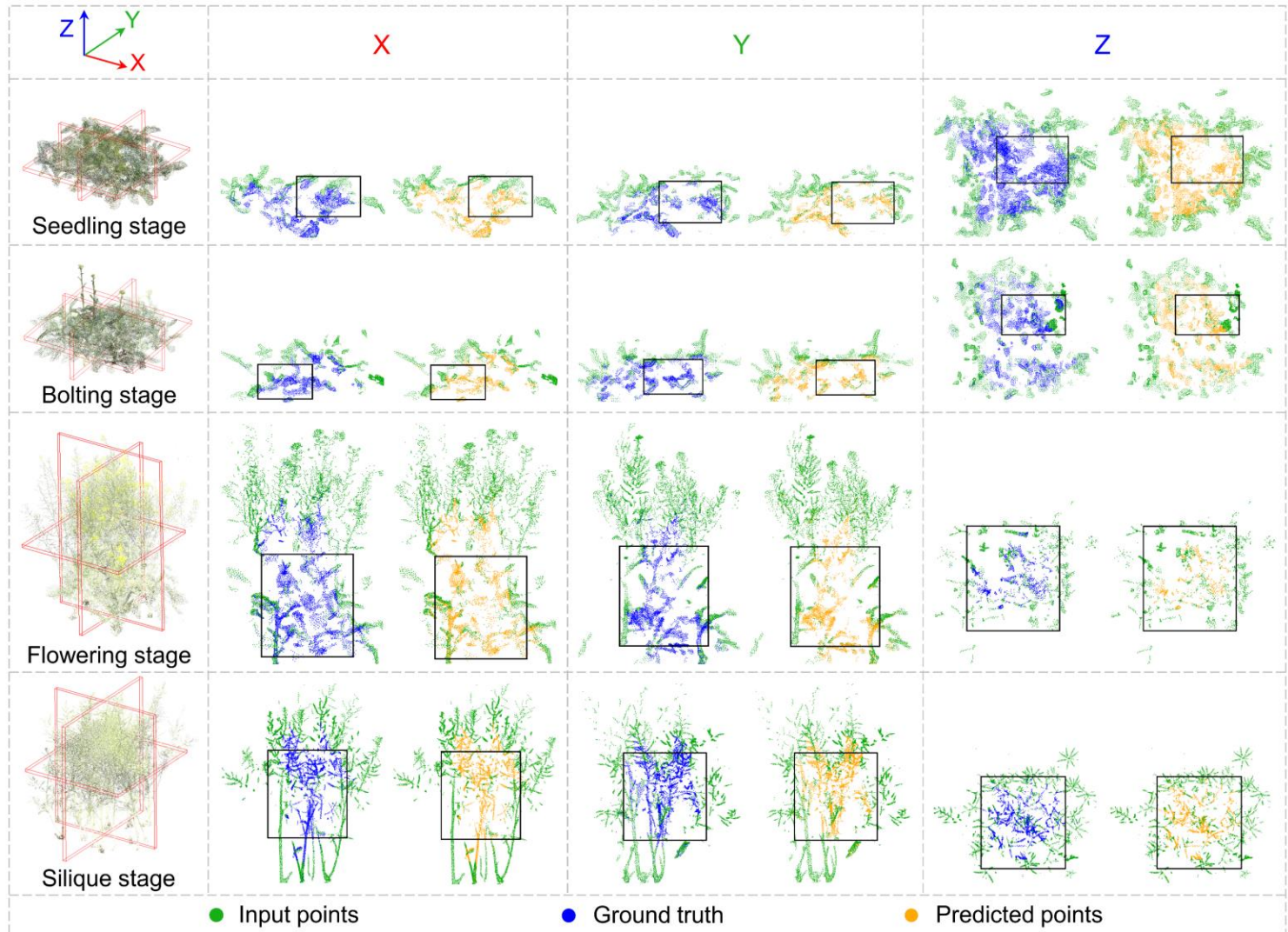

**Figure N1.** Visualization of CP-PCN point cloud completion results across different growth stages. The input points (green), ground truth (blue), and predicted points (orange) are shown for each stage: seedling, bolting, flowering, and silique. The black boxes highlight the magnified regions displayed in Figure 3, where the detailed comparisons between ground truth and predicted points are further illustrated.

## Supplemental Note 2. Processing time for model construction and application

Under the hardware and parameter conditions used in this study, re-training the model for a different crop takes approximately 150 hours (Table N1). While this time investment may seem lengthy, it is a one-time process for each crop, making it a reasonable investment for developing a robust, crop-specific model. Once trained, the model can be applied to large-scale phenotyping without requiring further modifications. When applying a pre-trained model for rapeseed point cloud completion, processing a single field plot takes approximately 350 seconds. The majority of this time is spent on pose estimation and 3D reconstruction, while the actual inference of missing point clouds using CP-PCN requires only a few seconds. Despite the overall processing time per plot being several minutes, this duration is acceptable for high-throughput phenotyping applications, as real-time processing is typically not a strict requirement. In practical field experiments, point cloud completion is generally performed as a post-processing step after data collection, which aligns well with real-world agricultural workflows.

Table N1. Processing time for model construction and application phases

|  | Step | Time cost |
|---|---|---|
| Model Construction | Pose estimation | 300 seconds / 150 images |
|  | NeRF training | 180 seconds |
|  | Building point cloud completion data set | 120 seconds / plot |

|  | Training point cloud completion model | 15 hours |
|---|---|---|
| Model Application | Pose estimation | 100 seconds / 36 images |
|  | NeRF training | 180 seconds |
|  | Model inference | 2 seconds |

**Supplemental Note 3. The detailed architecture of CP-PCN**

**Multi-resolution encoder**

To accommodate the size variation and architectural complexity of rapeseed populations across different growth stages, a multi-resolution feature extraction strategy was conducted in this study. The input point cloud was down-sampled using iterative farthest point sampling (IFPS) to three resolutions, containing $N$ (8192), $N/2$ (4096), and $N/4$ (2048) points, respectively. This hierarchical down-sampling mechanism enabled the network to capture both local and global features, ensuring that architectural details were preserved at multiple scales.

At each resolution level, the dynamic graph convolutional feature extractor (DGCFE) was applied through five layers, where each layer dynamically constructed local graphs to learn spatial relationships and high-level semantic features. Specifically, the feature dimensions of the extracted representations increased progressively from 64, 128, 256, 512, to 1024. To enhance feature aggregation, the output from the last four layers underwent a max pooling operation at each resolution, extracting the most salient feature representations. The pooled feature vectors were then concatenated across all layers, forming a 1920-dimensional latent vector ($F$). This comprehensive feature representation was subsequently aggregated into a final feature map ($M$), which integrated information from all resolutions to facilitate effective downstream point cloud completion.

**Dynamic graph convolutional feature extractor**

In the CP-PCN model, we designed a DGCFE to extract features from point clouds, leveraging its unique capability to dynamically update neighborhood relationships at each convolutional layer. Specifically, unlike traditional static graph architectures, the DGCFE recalculated k-nearest neighbor (k-NN) relationships based on the evolving feature space at each layer, allowing the network to flexibly adapt to variations in local geometry across layers and enhance spatial feature representation, as shown in Figure 7C.

The DGCFE mechanism was built around the EdgeConv operation, which constructed edge features by considering both the center point and its neighbors. For a given center point $p_i$ and one of its neighboring points $p_j$, EdgeConv computed an edge feature as follows:

$$h_{ij} = MLP([x_i, x_j - x_i]) \tag{1}$$

where $x_i$ and $x_j$ denoted the feature vectors of points $p_i$ and $p_j$, respectively, and $[x_i, x_j - x_i]$ represented the concatenation of the center point's features with the relative position vector. A MLP was then applied to transform these concatenated features, capturing complex spatial relationships within each neighborhood.

After computing edge features $h_{ij}$ for each point and its neighbors, a max pooling operation aggregated the edge features, resulting in a unique representation $h_i$ for the center point $p_i$:

$$h_i = max_{j=1,\ldots,k}(h_{ij}) \tag{2}$$

This pooling operation emphasized the most significant features within each neighborhood, ensuring that the representation remained invariant to the ordering of neighboring points. By repeating this process across multiple layers, the DGCFE progressively refined the neighborhood graph and captured detailed local structures in the point cloud data, yielding a robust and adaptive spatial feature representation. In this study, the size of $k$ was set to 20.

**Point pyramid decoder**

The point pyramid decoder (PPD) was a crucial component of the CP-PCN model, responsible for decoding the high-dimensional feature vector into the missing point cloud. Operating as a hierarchical decoder, the PPD processed the input feature vector at multiple resolutions, ultimately reconstructing the occluded point cloud at three distinct scales. The PPD took the 5760-dimensional final feature vector $V$ as input and generated the missing point cloud of size $M \times 3$, representing the shape of the missing region. The PPD operated as a hierarchical decoder, processing features at different resolutions. The input feature vector was passed through multiple stages to progressively refine the point cloud completion. Initially, the feature vector $V$ was passed through the fully connected layers, resulting in three feature layers: $FC_3$ (512 neurons), $FC_2$ (1024 neurons), $FC_1$ (1920 neurons). Each of these layers was responsible for predicting point cloud structures at different resolutions. At the deepest layer, $FC_3$ predicted the primary center points $Y_{coarse}$, which had a size of $M_1 \times 3$. These primary center points served as the foundation for further refinement. Then, $FC_2$ predicted the relative coordinates of secondary center points $Y_{middle}$. Each point in $Y_{coarse}$ acted as a center point to generate $M_2$ points of $Y_{middle}$, refining the structure of the predicted point cloud. The process involved "Expand" and "Add" operations, where the size of $Y_{middle}$ became $M_2 \times 3$. Finally, the detailed point cloud $Y_{fine}$ was predicted by $FC_1$ resulting in the final high-resolution prediction. The generation of $Y_{fine}$ followed the same principle as $Y_{middle}$ but focused on finer structural details, as illustrated in Fig. 2b. The size of $Y_{fine}$ is $M \times 3$, and it was designed to closely match the feature points sampled from the ground truth. By leveraging this hierarchical structure, high-level features extracted from the initial coarse layers influenced the generation of detailed features, thereby minimizing distortion and preserving the fine geometric details of the original missing point cloud. Here, the values of $M$, $M_1$, $M_2$ were set to 8192, 2048, 4096, respectively.

**Supplemental Note 4. Design of loss function**

The loss function played an essential role in training the CP-PCN model, guiding the optimization of both the feature extraction and point cloud completion processes. It consisted of two key components: multi-stage completion loss and adversarial loss. The completion loss measured the difference between the ground truth missing point cloud $Y_{GT}$ and the predicted point cloud, while the adversarial loss optimized the MRDG and PPD to ensure that the predicted output appeared more realistic. The size of $Y_{GT}$ was $M \times 3$, which matched $Y_{fine}$. The commonly used chamfer distance (CD), widely applied in point cloud completion research, was adopted as the model's loss function, which was calculated as follows:

$$d_{CD}(S_1, S_2) = \frac{1}{S_1} \sum_{x \in S_1} \min_{y \in S_2} \|x - y\|_2^2 + \frac{1}{S_2} \sum_{y \in S_2} \min_{x \in S_1} \|y - x\|_2^2 \qquad (3)$$

where $S_1$ and $S_2$ represented two sets of 3D point clouds. CD in Equation (3) measured the average nearest squared distance between the predicted point cloud $S_1$ and the ground truth point cloud $S_2$. Since the PPD predicted three point clouds at different resolutions, the multi-stage completion loss was formulated as follows:

$$L_{com} = d_{CD1}(Y_{fine}, Y_{GT}) + \alpha\, d_{CD2}(Y_{middle}, Y'_{GT}) + 2\alpha\, d_{CD3}(Y_{coarse}, Y''_{GT}) \qquad (4)$$

where $d_{CD1}$, $d_{CD2}$ and $d_{CD3}$ represented the CD at different resolution levels, weighted by the hyperparameter $\alpha$. The weighting parameter $\alpha$ was progressively updated as the number of training epochs increased. The first term calculated the squared distance between the detailed points $Y_{fine}$ and the ground truth of the missing region $Y_{GT}$. The second and third terms calculated the squared distance between the primary center points $Y_{coarse}$ and secondary center points $Y_{middle}$ and their corresponding subsampled ground truth $Y''_{GT}$, $Y'_{GT}$, respectively. The subsampled ground truth $Y''_{GT}$ and $Y'_{GT}$ were obtained by applying IFPS to extract representative feature points from the missing region and were of sizes $M_1 \times 3$ and $M_2 \times 3$, respectively.

The adversarial loss was inspired by GANs. The generator function was defined as $F(\ ) = PPD(MRDG())$, where $F: \mathcal{X} \to \mathcal{Y}'$ mapped the partial input $\mathcal{X}$ to the predicted missing region $\mathcal{Y}'$. The discriminator $D()$ attempted to distinguish between the predicted missing region $\mathcal{Y}'$ and the real missing region $\mathcal{Y}$. The discriminator incorporated both a DGCFE module and an MLP network to compare the predicted point clouds with the ground truth. The structure of the DGCFE remained consistent with its implementation in the MRDG. The MLP network consisted of sequential layers with dimensions [64 – 64 – 128 – 256], where max pooling was applied to the outputs from the last three layers to obtain a compact feature representation. These pooled feature vectors were concatenated to form a 448-dimensional latent vector, which was subsequently passed through fully connected layers [256, 128, 16, 1], and a sigmoid classifier to generate the final discrimination result. The adversarial loss was computed based on the discriminator's performance as follows:

$$L_{adv} = \sum_{1 \leq i \leq S} \log(D(y_i)) + \sum_{1 \leq i \leq S} \log\left(1 - D(F(x_i))\right) \tag{5}$$

where $x_i \in \mathcal{X}$, $y_i \in \mathcal{Y}, i = 1, 2, \ldots, S$, with $S$ representing the dataset size. Both $F()$ and $D()$ were jointly optimized using alternating Adam updates during training.

The final loss function, combining the multi-resolution completion loss and adversarial loss, was formulated as:

$$L = \lambda_{com}L_{com} + \lambda_{adv}L_{adv} \tag{6}$$

where $\lambda_{com}$ and $\lambda_{adv}$ were weight hyperparameters satisfying $\lambda_{com} + \lambda_{adv} = 1$. In this study, $\lambda_{com}$ was set to 0.9.